\documentclass[runningheads, a4paper]{llncs}
\pdfoutput=1

\usepackage[hidelinks]{hyperref}

\usepackage{graphicx}
\usepackage{todonotes}
\usepackage[inline]{enumitem}

\usepackage[nohyperlinks,nolist]{acronym}
\usepackage{eso-pic}
\usepackage{xspace}
\usepackage[binary-units=true]{siunitx}
\usepackage{multirow}
\usepackage{subcaption}
\usepackage{booktabs}

\makeatletter
\DeclareRobustCommand\onedot{\futurelet\@let@token\@onedot}
\def\@onedot{\ifx\@let@token.\else.\null\fi\xspace}
\def\eg{\emph{e.g}\onedot} 
\def\ie{\emph{i.e}\onedot}

\def\etal{\emph{et al}\onedot}

\newcommand{\printfnsymbol}[1]{%
  \textsuperscript{\@fnsymbol{#1}}%
}

\newcommand{\ganwriting}{GANwriting\xspace}

\begin{document}
\title{Handwriting Classification for the Analysis of Art-Historical Documents}

\begin{acronym}
	\acro{WPI}{Wildenstein Plattner Institute}
	\acro{OCR}{Optical Character Recognition}
	\acro{IAMDB}{IAM Handwriting Database}
	\acro{IAM-HistDB}{IAM Historical Handwriting Database}
	\acro{IAMONDB}{IAM On-line Handwriting Database}
	\acro{cGAN}{conditional Generative Adversarial Network}
	\acro{AdaIN}{Adaptive Instance Normalization}
	\acro{llr}{Log-Likelihood Ratio}
	\acro{5CHPT}{Five Classes Handwritten and Printed Text}
	\acro{PCA}{Principal Component Analysis}
\end{acronym}

%
%
\author{Christian Bartz\thanks{equal contribution} \and
Hendrik R\"atz\printfnsymbol{1} \and
Christoph Meinel}
\authorrunning{C. Bartz, H. R\"atz, et al.}
%
\institute{Hasso Plattner Institute \\ University of Potsdam \\ 14482 Potsdam Germany \\
\email{[christian.bartz,christoph.meinel]@hpi.de},
\email{hendrik.raetz@student.hpi.de}
}
\maketitle              
\begin{abstract}
Digitized archives contain and preserve the knowledge of generations of scholars in millions of documents.
The size of these archives calls for automatic analysis since a manual analysis by specialists is often too expensive.
In this paper, we focus on the analysis of handwriting in scanned documents from the art-historic archive of the \acl{WPI}.
Since the archive consists of documents written in several languages and lacks annotated training data for the creation of recognition models, we propose the task of handwriting classification as a new step for a handwriting \acs{OCR} pipeline.
We propose a handwriting classification model that labels extracted text fragments, \eg, numbers, dates, or words, based on their visual structure.
Such a classification supports historians by highlighting documents that contain a specific class of text without the need to read the entire content.
To this end, we develop and compare several deep learning-based models for text classification.
In extensive experiments, we show the advantages and disadvantages of our proposed approach and discuss possible usage scenarios on a real-world dataset.

\end{abstract}

\section{Introduction}

Archives contain a wealth of documents that are highly valuable for historical research since they have been gathered over a long period of time (typically several centuries or decades).
Digitization helps to preserve documents, which only exist on paper, by converting the documents into a format that can be copied and distributed without quality loss.
However, simply digitizing documents does not provide additional value for research.
The digitized documents are only available in the form of digital images, thus preventing a direct search of their content.
Data that is of interest includes, but is not limited to, the content of printed texts, handwritten texts, and depicted objects in images.
The extraction of such metadata is a time and labor-intensive task, which can be done by humans, but becomes infeasible very quickly once an archive contains millions of documents.

Thanks to recent advances in machine learning and computer vision~\cite{he_mask_2017,krizhevsky_imagenet_2012,wigington_start_2018}, it is possible to automate the task of analyzing digitized data of an archive.
In our work, we focus on the analysis of handwriting because it may contain thoughts or the contents of personal communication.
Additionally, it might provide extra information that can not be found elsewhere, such as prices of artworks at an auction.
In our work, we analyze the digitized data of the art-historical archive of the \ac{WPI}\footnote{\url{https://wpi.art}}, whose archive contains information that is, \eg, of interest when researching the provenance of works of art.
\ac{OCR} is the tool of choice to recognize handwritten text contained in the digitized documents.
In a typical setting, an \ac{OCR} system first locates all regions that contain printed or handwritten text and then it recognizes the content of words, groups of words, or entire lines of text.

Modern approaches for \ac{OCR} make use of deep learning methods for successful recognition of text~\cite{ul-hasan_high-performance_2013,wigington_start_2018} (more on related work can be found in \autoref{sec:related_work}).
However, methods based on deep learning require a large amount of annotated training data for the language the model is to be applied to.
The availability of such data is crucial for the creation of modern deep-learning-based handwriting recognition models.
Therefore, it is not directly possible to apply research results for handwriting recognition~\cite{kang_convolve_2019,wigington_start_2018} on the archive of the \ac{WPI} because no annotated data is available.
To extract information without fully recognizing the textual content of a given text snippet, we propose to classify images of text based on their visual structure.
Such a classification provides researchers with information about the type of text in their documents, \eg, a page does not only contain words but also dates or other identifiers.
This helps them to identify and filter documents and pages quickly by specifying the type of information they are looking for.

In this paper, we introduce and evaluate several approaches for text classification.
On the one hand, we propose to train a softmax classifier with a fixed number of classes.
On the other hand, we propose to learn a model that uses metric learning to embed images of handwriting.
This model arranges the images in a way that those with similar structure are located close to each other, while positioning structurally different images far from each other (more on our models can be found in \autoref{sec:method}).
In our experiments (see \autoref{sec:experiments}), we show that our models perform well on different datasets.
The softmax classifier outperforms the metric learning approach, but the metric learning approach is by far more flexible and can be used in a more general setting.
The contributions made in this paper can be summarized as follows:
\begin{enumerate*}[label={\arabic*)}]
	\item An in-depth analysis of handwriting classification for the analysis of archival data.
	\item Introduction of methods for successfully synthesizing training data, such as numbers or dates, which are not or rarely contained in existing datasets for handwriting recognition.
	\item We provide our trained models and our code to the research community for further experimentation\footnote{\url{https://github.com/hendraet/handwriting-classification}}.
\end{enumerate*}

\section{Related Work}
\label{sec:related_work}

Handwriting recognition is a field of \ac{OCR} that has actively been researched over the past decades~\cite{tappert_state_1990,plamondon_online_2000,bhunia_handwriting_2019,wigington_data_2017,wigington_start_2018,carbune_fast_2020}.
In this section, we will introduce work from the area of handwriting recognition and also existing work in the area of handwriting classification.

\paragraph{Handwriting Recognition}
Online handwriting recognition describes the task of recognizing handwritten characters based on the recorded movement information of a person writing characters.
The field of online handwriting recognition has been researched very actively in the past and is still of interest to the community~\cite{tappert_state_1990,plamondon_online_2000,carbune_fast_2020}.
However, in the area of archival analysis, we do not have access to online writing data because the handwriting data is only available in image form.

Offline handwriting recognition revolves around the task of recognizing handwriting in images of text.
Over time, approaches shifted from using Hidden Markov Models~\cite{gimenez_handwriting_2014,bluche_tandem_2013} to the usage of deep neural networks in conjunction with recurrent neural networks~\cite{wigington_start_2018,bhunia_handwriting_2019,wigington_data_2017,kang_convolve_2019}.
The proposed models reach state-of-the-art results with character errors rates as low as \SI{8}{\percent}~\cite{bhunia_handwriting_2019} on the IAMDB dataset~\cite{marti_iam-database_2002}.
However, these results can only be achieved for handwriting of a specific language (English in the case of the IAM dataset) while using a large amount of data for the training of the recognition model.
When trying to use a system on another language, such as French or Italian, an entirely new model, which requires a large amount of annotated training data, needs to be created.
This problem gets even more severe when dealing with documents containing older scripts or words that are not used anymore.

\paragraph{Handwriting Classification}
To alleviate the problem of missing training data and to allow at least some insight into the data, classifying text snippets based on their visual structure could be a viable alternative.
To the best of our knowledge, not much research has been done in that direction.
Mandal~\etal~\cite{mandal_multi-lingual_2015} propose a method that takes text snippets as input and locates dates within documents.
In their paper, they try to find month words, such as ``Jan'' or ``October'', using handcrafted features.
Additionally, they identify digits using gradient-based features in conjunction with a Support Vector Machine.


\section{Method}
\label{sec:method}

The task of handwriting classification enables us to analyze handwriting without the need to directly recognize textual content.
This enables analysis of handwritten documents from different languages by looking at the structure of the depicted handwriting.
With our models, we try to discern handwritten words from numbers, dates, and also alphanumeric strings.
Existing handwriting databases (\eg, the \ac{IAMDB}~\cite{marti_iam-database_2002}) contain a large amount of handwritten text but hardly any numbers, dates, or alphanumeric strings.
In this section, we first introduce a strategy to synthesize enough data for the training of our models.
Then, we introduce our first model, which is a simple softmax classifier.
Additionally, we train an embedding model based on metric learning since we hypothesize that such a model is more flexible than a softmax classifier with a fixed number of classes.

\subsection{Handwriting Synthesis}
\label{subsec:handwriting_synthesis}

Existing datasets, such as the well-known IAM datasets (\ac{IAMDB}~\cite{marti_iam-database_2002}, \ac{IAM-HistDB}~\cite{fischer_lexicon-free_2012,fischer_transcription_2011}), rarely contain numbers, dates, or alphanumeric strings.
The largest of the IAM datasets, the \ac{IAMDB} dataset, contains \num{115320} images of words but only 454 images where numbers are displayed.
The unavailability of numbers, dates, and alphanumeric strings poses a problem for our approach.
A solution to this problem is the usage of synthetic but realistically looking handwriting.
Approaches for the generation of synthetic handwriting exist, but synthetic datasets, such as the IIIT-HWS~\cite{krishnan_generating_2016} dataset, also only contain a small amount of samples that are not words.

\begin{figure}[t]
  \centering
  \includegraphics[width=\textwidth]{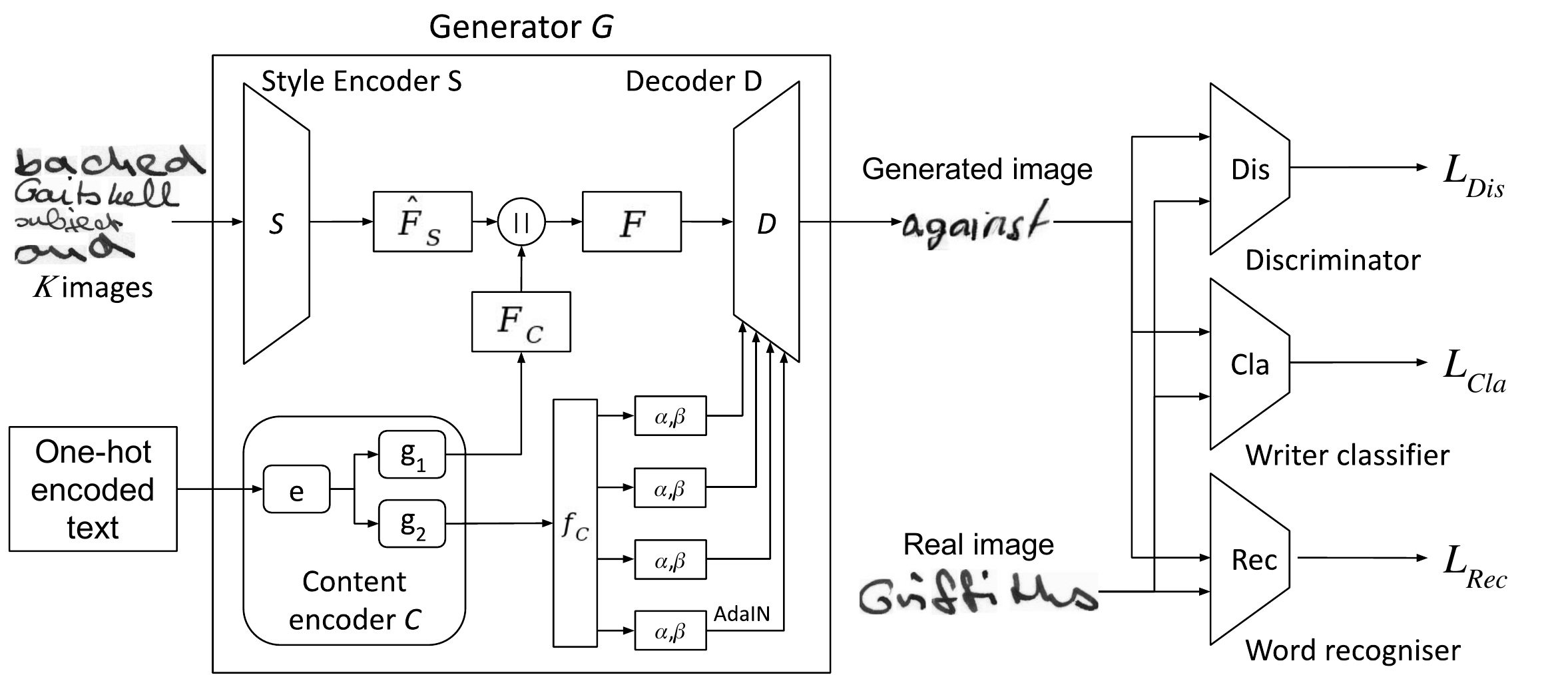}
  \caption{
    Structural overview of the \ganwriting generator model.
    The model consists of a generator that receives the text to be generated and a set of \emph{K} images that define the desired style.
    The generated images are assessed by three models: a discriminator, a writer classifier, and a word recognizer.
  }
  \label{fig:ganwriting_model_overview}
\end{figure}

To overcome the problem of missing handwritten numbers and dates, we adapt the \emph{\ganwriting} model proposed by Kang~\etal~\cite{kang_ganwriting_2020}.
The \ganwriting model is based on a \ac{cGAN} that produces realistic looking handwriting conditioned on input styles.
Furthermore, the model is able to synthesize words or character combinations that have not been part of the train set.
A structural overview of the model can be seen in \autoref{fig:ganwriting_model_overview}.
In the following, we describe the most important parts of the model and our changes to it.

\paragraph{Generator}
The generator of the \ganwriting model takes two inputs.
The first set of inputs are multiple images depicting handwritten words of one author.
These inputs are passed to the style encoder, which extracts the stylistic features of the handwriting.

The second input is a vector with a one-hot-encoded representation of the input characters.
This input is passed through the content encoder, generating two outputs.
The first output is a character-wise encoding that is concatenated with the extracted stylistic features, which are subsequently used to steer the generation of the characters in the resulting image.
Here, we adapted the model and increased the maximum allowable word length from \textbf{seven} characters to a maximum of \textbf{twenty five} characters, by introducing a novel padding strategy.
In a second step, the content encoder generates parameters that are used by the decoder to perform \ac{AdaIN}~\cite{huang_arbitrary_2017}.

Style and content features are used to guide the decoder model in the generation of handwriting, which is rendered in the style of the input images while containing all required characters.

\paragraph{Discriminator}
In the \ganwriting model, not one but three models are used for assessing the quality of the generated image.
The first model is a discriminator that tries to determine whether a given input image is fake or real.
The second model is a writer classifier, whose task is to determine the author of a given handwriting sample.
Thus, it forces the generator to make use of the stylistic information passed to the style encoder and asserts that synthesized images use the provided style.
The third model is a text recognition model, which is used to force the generator to create handwriting that is readable while containing the expected content.
For further information about the three models used to assess the quality of the generator, please refer to the \ganwriting paper~\cite{kang_ganwriting_2020}.

\paragraph{Data Generation with the \ganwriting Model}
The described model can be used to generate new handwriting images, to generate handwritten numbers, and dates, we still need some examples of these types of data.
Since the available datasets do not contain enough of these samples, we used another mechanism to synthesize simple numbers and dates that we use for training of our adapted \ganwriting model.
To produce these samples, we make use of online handwriting data from the \ac{IAMONDB}~\cite{liwicki_iam-ondb_2005}.
Here, we manually extract the strokes of individual digits, as well as the strokes for dots and dashes.
We then generate numbers and dates by concatenating the gathered strokes and rendering them on a white canvas.
It is important to note that we can not combine the strokes of different authors because the \ganwriting model learns to mimic the style of one author per generated sample.
Using synthesized data with strokes from different authors would, therefore, lead to unsatisfactory results.
The data synthesized in that way does not contain any information about stroke thickness or color of the stroke because online data only contains information about movements while writing.
Thus, this information is artificially generated and added before the rendering process to simulate a broader variety of writing styles.
We then use the newly generated dates and numbers together with the already existing word images from the \ac{IAMDB} dataset for the training of the \ganwriting model.

\subsection{Handwriting Classification Networks}
\label{subsec:handwriting_classification_networks}

Besides data, we need classification approaches for categorizing given handwritten words.
In the following, we introduce two different architectures for the classification of handwriting.
On the one hand, we propose the usage of a softmax-based classifier.
On the other hand, we propose to use a distance-based model that can be used to embed images in a way that the Euclidean distance can be used as a similarity measure.
Using such an embedding model, we can determine clusters and classify new samples based on their location in these clusters.

\paragraph{Softmax Classifier}
Our first model is a ResNet-18~\cite{he_deep_2016} based network with a softmax classifier, which can discern the classes present in our experiments (see \autoref{sec:experiments}).
We train the model using softmax cross entropy as loss function.
Classifiers trained with softmax cross entropy are very strong multi-class classification models, but they are quite inflexible.
For instance, it is not possible to simply add a new class to the classifier without retraining the whole network.
To mitigate this issue, we propose another approach for text classification.

\paragraph{Distance Based Model}
In our distance-based model we also make use of a ResNet-18 feature extractor.
However, we increase the number of the neurons of the final linear classification layer and omit the softmax activation.
Therefore, the model transforms an input image into an embedding, which is is a vector with \num{512} dimensions.
We train our model to produce embeddings that are similar to each other if the input images depict the same type of content (\eg, words, dates, numbers, $\ldots$).
To learn the parameters of the model we make use of the triplet loss function~\cite{schroff_facenet_2015}.
When training a model with triplet loss, the model takes a triplet of samples as input.
The three samples are called \emph{anchor}, \emph{positive}, and \emph{negative}, where anchor and positive belong to the same class and negative represents a different class.
The objective of a model trained using triplet loss is to minimize the Euclidean distance between the anchor and positive sample, while maximizing the Euclidean distance between the anchor and the negative sample.
Effectively, the model groups samples of the same class, while it positions those of different classes far from each other.

Based on a properly trained triplet loss model, it is possible to group embeddings of samples in clusters based on their class.
A sample can now be classified using multiple strategies.
The first strategy (we refer to this as the \emph{naive} approach) is to use two widely used algorithms: k-means~\cite{lloyd_least_1982} for clustering and kNN~\cite{altman_introduction_1992} for classification.
We first cluster the embeddings of unseen samples and determine the centroids of the found clusters.
To fit the kNN model, we then use a set of support embeddings, \ie, embeddings where the class is known, and determine the class of the clusters based on the embedded support samples.

Another possible approach for classification is to use distance-based thresholding~\cite{schroff_facenet_2015}.
In distance-based thresholding the distances of the embeddings of a questioned sample and a support sample are compared.
If the distance of the questioned sample to the support sample is greater than a predefined threshold, the sample is classified as not belonging to the support sample's class.
However, if the distance is lower than the predefined threshold, the sample is classified as belonging to the class of the support sample.
Using such a \emph{hard decision} threshold is impractical, because the threshold needs to be calculated for each model and each class individually, rendering the proposed distance-based model inflexible when handling new and unseen data.

A better approach is to produce \emph{soft decisions} based on \acp{llr}~\cite{brummer_application-independent_2006}.
Opposed to the hard decision based on a threshold, \acp{llr} provide a relative score reflecting a confidence.
The idea is that the \ac{llr} determines how likely it is that a calculated distance \emph{d} is observed for a sample of class \emph{A} (target trial) or class \emph{B} (non-target trial).
To determine the likelihood, it is necessary to compute the distance distributions for the two classes beforehand by using a validation/support set.
The logarithm of the ratio of those likelihoods then forms the llr.
The \ac{llr} was designed with the application of two-class classification in mind.
However, they can also be used in a multi-class setting, where the \ac{llr} is calculated in a one-vs-all approach for all classes.

The distance-based approach is very flexible.
If the embedding network learns to extract meaningful features that describe the structure of the underlying text, it is theoretically possible to classify samples of classes unseen during training
Such a behavior is not possible when using a softmax-based classifier, since the classifier would need to be retrained for each new class, which means that enough training data needs to be gathered.


\section{Results and Discussion}
\label{sec:experiments}

In this section, we present the results of our experiments on multiple datasets.
We show that our proposed models are feasible for usage on synthetic and real data.
Furthermore, we discuss the advantages of our proposed models, especially their generalization capabilities and their ability to handle classes that have not be seen during training.
First, we introduce the datasets we used for our experiments.
Subsequently, we provide some information about our experimental setup.
Last, we present the results of our experiments in addition to the insights we draw from them.

\subsection{Datasets}

\begin{figure}[t]
  \centering
  \includegraphics{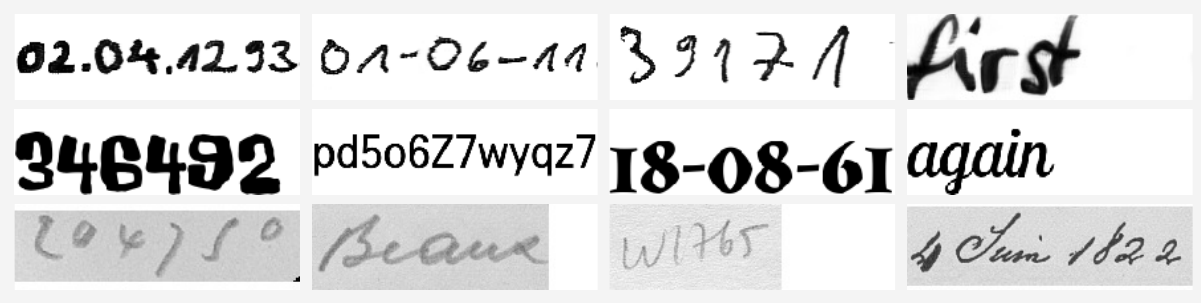}
  \caption{
    Visualization of samples from all datasets used in this work.
    In the first row, we show samples from the \ganwriting dataset.
    The second row depicts (printed) samples, which, together with samples from the \ganwriting dataset, form the \acs{5CHPT} dataset.
    The last row shows samples from the \ac{WPI} dataset, which solely consists of real images from archival documents and is used for evaluation only.
  }
  \label{fig:datasets}
\end{figure}

In our experiments, we use multiple datasets for the evaluation of our proposed models.
These datasets contain different kinds of samples distributed over various classes.
On the one hand, we use two different synthetic datasets for training and preliminary evaluation.
On the other hand, we employ a dataset consisting of real samples from the archive of the \acl{WPI} for evaluation purposes only.
We show some samples from each dataset in \autoref{fig:datasets}.

\paragraph{Synthetic Datasets}

As already mentioned in \autoref{subsec:handwriting_synthesis}, getting access to a large amount of labelled training data is very difficult, if not impossible.
Hence, we devise methods to synthesize data that we can use for the training of our proposed models.
Here, we create two different datasets.
The first dataset consists of samples synthesized by our adaption of the \ganwriting model (see \autoref{subsec:handwriting_synthesis}).
This dataset consists of synthetic handwriting samples from three classes (words, numbers and dates), totalling in \num{7920} samples that are equally distributed over the classes.
The second dataset adds two more classes: alphanumeric strings and five-digit numbers that resemble zip codes.
In addition to synthetic handwritten text, this dataset includes samples of synthetic printed text.
We decided to also add printed text, as the model should be trained for classifying images of text, based on the visual structure of the contained texts.
We argue that adding printed text leads to extra variety in our samples, which helps our model to generalize better because the structure of handwritten text and printed text for, \eg, dates, is similar.
Since the dataset consists of 5 classes and contains handwritten, as well as printed text, we refer to it as \ac{5CHPT} dataset.
In total, the dataset consists of \num{26400} samples, equally distributed over the five classes.

\paragraph{WPI Dataset}

Besides the synthetic datasets that we use for training and validation of our models, we also gathered a dataset from the archive of the \ac{WPI}.
This dataset consists of \num{272} images of handwritten text, showing numbers (\num{112} samples), words (\num{108} samples), alphanumeric string (\num{31} samples), and dates (\num{21} samples).
We depict some of the samples in the last row of \autoref{fig:datasets}.
Please note, we first binarize images from this dataset before feeding them to our models.

\begin{table}[t]
    \caption{
        Evaluation results for our proposed models on all of our datasets.
        We report accuracy, precision, recall, and F1 score.
        \textbf{Bold font} indicates the best result on the corresponding dataset.
    }
    \label{tab:results_ganwriting}
    \begin{center}
        \begin{tabular}{@{}l c c c c c@{}}
    \toprule
    Dataset                         & Model  & Accuracy (\SI{}{\percent})   & Precision (\SI{}{\percent})   & Recall (\SI{}{\percent})  & F1 Score         \\
    \midrule
    \multirow{3}{*}{\ganwriting}    & Naive  & \textbf{99.62}               & \textbf{99.62}                & \textbf{99.62}            & \textbf{0.9962}  \\
                                    & LLR    & 99.24                        & 99.25                         & 99.24                     & 0.9924           \\
                                    & Softmax& \textbf{99.62}               & \textbf{99.62}                & \textbf{99.62}            & \textbf{0.9962}  \\
    \midrule
    \multirow{3}{*}{\ac{5CHPT}}     & Naive  & 74.32                        & 73.31                         & 74.32                     & 0.7456           \\
                                    & LLR    & 79.89                        & 99.25                         & 99.24                     & 0.7999           \\
                                    & Softmax& \textbf{90.00}               & \textbf{90.62}                & \textbf{90.00}            & \textbf{0.9004}  \\
    \midrule
    \multirow{3}{*}{\ac{WPI}}       & Naive  & \textbf{58.33}               & 49.82                         & \textbf{58.33}            & \textbf{0.5263}  \\
                                    & LLR    & 25.76                        & 32.96                         & 25.76                     & 0.2771           \\
                                    & Softmax& 37.50                        & \textbf{60.61}                & 37.50                     & 0.4086           \\
    \bottomrule
    \end{tabular}
    \end{center}
\end{table}

\subsection{Experimental Setup}

The model we use to synthesize handwriting samples is based on Lei Kang's Pytorch implementation\footnote{\url{https://github.com/omni-us/research-GANwriting}} of the \ganwriting paper~\cite{kang_ganwriting_2020}.
Our adapted version that additionally can handle date generation is also freely available on GitHub\footnote{\url{https://github.com/hendraet/research-GANwriting/tree/support-date-generation}}.
We train our \ganwriting model in two rounds for \num{3000} and \num{6000} epochs, respectively, using a batch size of \num{16} and a GPU with a total of \SI{12}{\giga\byte} of RAM.
We use Adam~\cite{kingma_adam:_2015}, as our optimizer, with a learning rate of $10^{-4}$ for discriminator and generator and $10^{-5}$ for writer classifier and content recognizer.
We implement the classification models using the deep learning framework Chainer~\cite{tokui_chainer:_2015} and provide our code on GitHub\footnote{\url{https://github.com/hendraet/handwriting-classification}}.
Here, we also use a GPU with a total of \SI{12}{\giga\byte} of RAM, Adam with a learning rate of $10^{-4}$, and a batch size of \num{128}.
We train these models for \num{20} epochs.

\subsection{Results on the \ganwriting Dataset}

\begin{figure}[t]
    \centering
    \begin{subfigure}{0.49\textwidth}
         \centering
         \includegraphics[width=\textwidth]{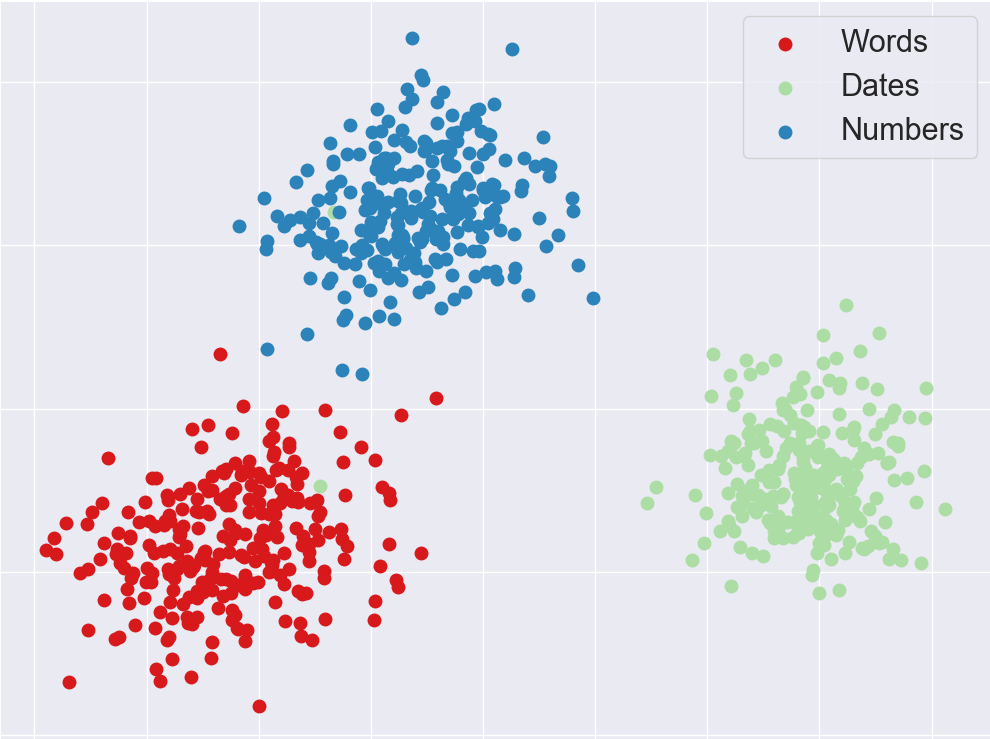}
         \caption{PCA of samples in \ganwriting dataset}
         \label{fig:pca_ganwriting}
    \end{subfigure}
    \hfill
    \begin{subfigure}{0.49\textwidth}
         \centering
         \includegraphics[width=\textwidth]{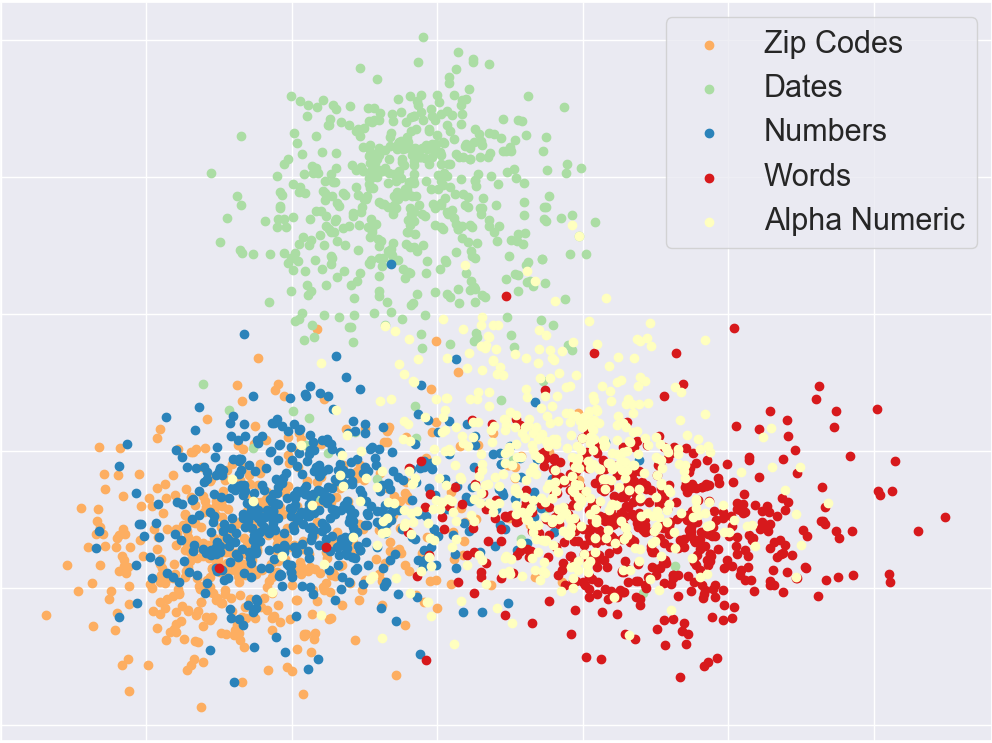}
         \caption{PCA of samples in \ac{5CHPT} dataset}
         \label{fig:pca_5chpt}
    \end{subfigure}
    \caption{
        Comparison of embeddings on our synthetic training datasets.
        Embeddings produced by the model trained on \ganwriting are shown in \subref{fig:pca_ganwriting}, whereas embeddings of the model trained on \ac{5CHPT} are shown in \subref{fig:pca_5chpt}.
        The visualizations are obtained by performing a \ac{PCA}~\cite{pearson_liii_1901}
    }
    \label{fig:pca_synthetic}
\end{figure}

In our first experiments, we assess whether our overall approach of classifying handwriting is feasible and test it on our \ganwriting dataset, which consists of images from three different classes.
The results (see the first rows of \autoref{tab:results_ganwriting}) show that all of our three approaches can correctly classify samples from the \ganwriting dataset.
When looking at a visualization of the predicted clusters (see \autoref{fig:pca_ganwriting}), we can see that the model is able to cluster the input samples into three distinct clusters.
Hence, the naive method relying on k-Means and kNN, as well as the \ac{llr} method perform very well.
We can also observe that the visual structure of dates differs significantly from the structure of numbers and words because the cluster of dates is positioned far away from the other clusters.

\subsection{Results on the \ac{5CHPT} Dataset}

In our next series of experiments, we trained our models on the \ac{5CHPT} dataset.
We examine whether our models can handle a wider variety of classes that are more similar to each other than the classes in the last experiment.
We present the results of these experiments in the second row of \autoref{tab:results_ganwriting}.
The results show that the model using a softmax classifier outperforms the embedding models by a large margin.

Consulting the visualization of the predicted clusters in \autoref{fig:pca_5chpt}, we can identify the reason for the comparatively poor performance of the embedding-based models.
Although individual clusters are visible, the clusters for zip codes, numbers, alphanumerics and words can not be distinguished clearly.
This is not unexpected considering the nature of the underlying classes.
Zip codes are a sub-class of numbers and alphanumerics are a superclass of words and numbers.
This complicates the clustering process for our embedding models, whereas the softmax model can rely on different features and push the samples into more distinct categories.
However, the embeddings of our date images form their own distinct cluster, showing that our embedding models indeed learn to cluster based on the structure of the given text sample.

The results on this dataset show the limitations of our classification approach.
Classification only makes sense for distinguishing visually dissimilar classes from each other in a coarse way.
A fine-grained classification is only possible with a full recognition pipeline with additional analysis steps, such as named entity recognition.

\subsection{Results on the \ac{WPI} Dataset}

\begin{figure}[t]
    \centering
    \begin{subfigure}{0.325\textwidth}
         \centering
         \includegraphics[width=\textwidth]{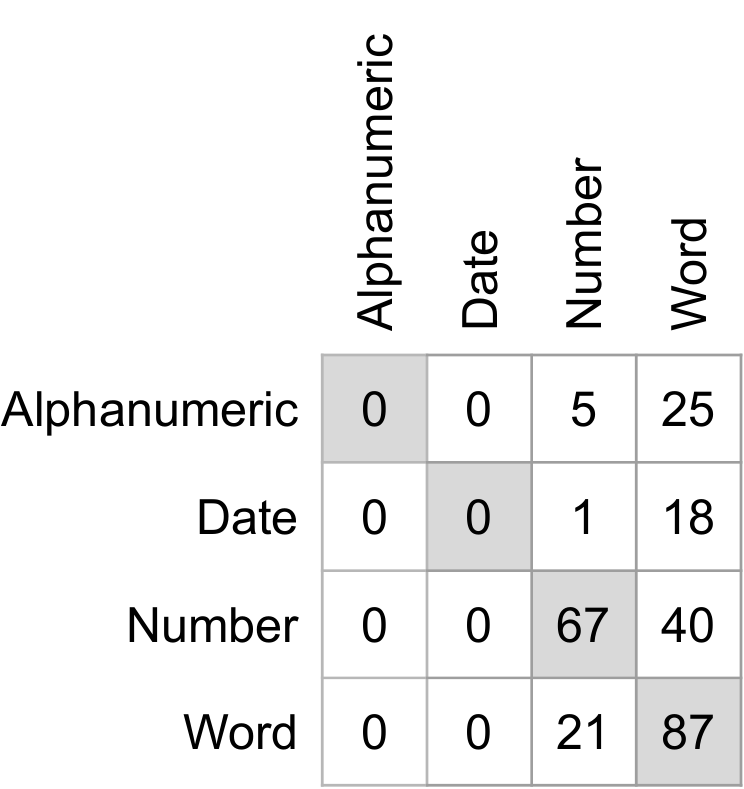}
         \caption{naive classifier}
         \label{fig:confusion_naive_approach}
    \end{subfigure}
    \hfill
    \begin{subfigure}{0.325\textwidth}
         \centering
         \includegraphics[width=\textwidth]{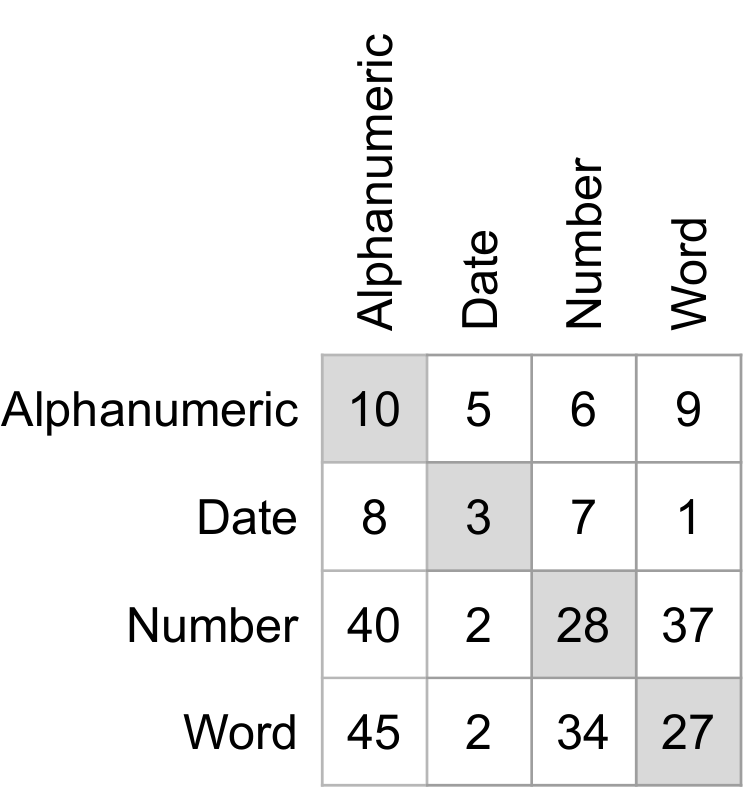}
         \caption{\ac{llr} classifier}
         \label{fig:confusion_llr}
    \end{subfigure}
    \hfill
    \begin{subfigure}{0.325\textwidth}
         \centering
         \includegraphics[width=\textwidth]{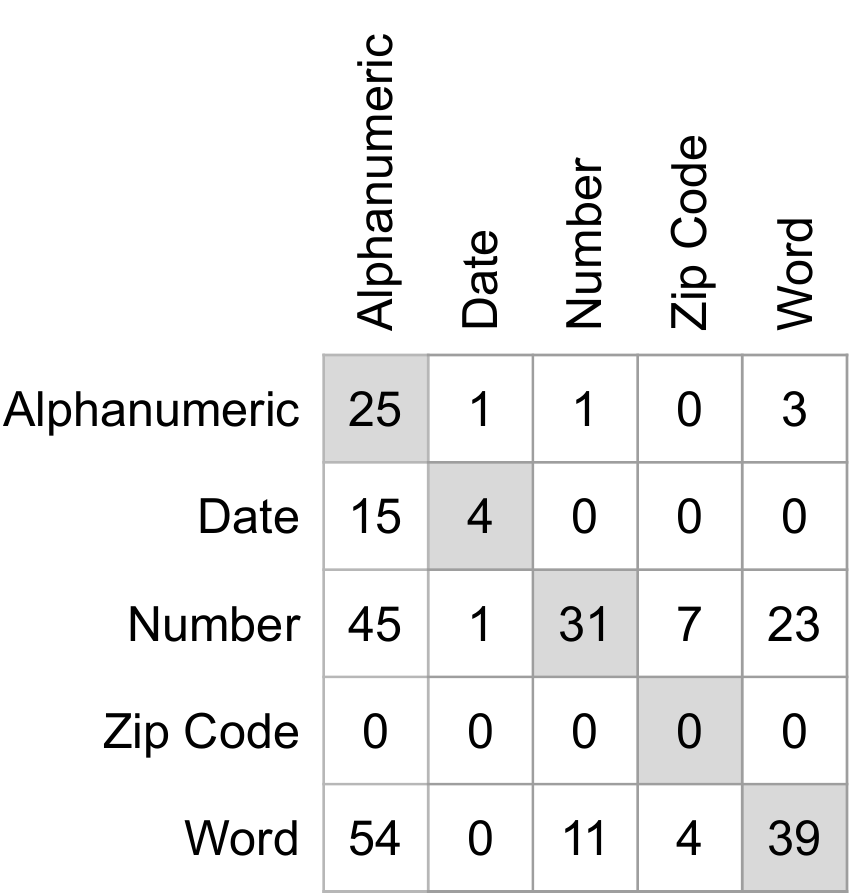}
         \caption{softmax classifier}
         \label{fig:confusion_softmax}
    \end{subfigure}
    \caption{
        Confusion matrices for our different classifiers on the \ac{WPI} dataset.
        Rows refer to the actual class, whereas columns refer to the predicted class.
        The confusion matrix of the softmax model also includes the zip code class because it was trained on the \ac{5CHPT} dataset and is therefore restricted to classifying exactly five classes.
    }
    \label{fig:wpi_dataset_confusion_pca}
\end{figure}

Following our experiments on the \ac{5CHPT} dataset, we evaluate the models trained on the \ac{5CHPT} dataset on the dataset that contains real images from the art-historical archive of the \ac{WPI}.
The results (see third row of \autoref{tab:results_ganwriting}) show that applying our trained models to the \ac{WPI} dataset leads to severe performance degradations.
Here, our naive model seems to perform best, which is an interesting result.

On closer examination the reasons for the seemingly better performance of the naive model become apparent.
Looking at at the confusion matrix of the naive approach (see \autoref{fig:confusion_naive_approach}), we can observe that the naive model is not able to identify alphanumerics and dates at all (F1 score of \num{0} for both classes).
The confusion matrix clearly shows that the naive classifier can only identify two classes.
We conclude, that the model is not able to embed the samples of the \ac{WPI} dataset so that they form distinct clusters.
Hence, the k-Means algorithm is only able to find two clusters: one for words and one for numbers.
This leads to a superior accuracy because the dataset is imbalanced and those two classes contain a majority of the samples.
Therefore, the negative impact of misclassifying dates or alphanumeric strings is comparatively small.
The \ac{llr} classifier, on the other hand, is able to find at least some instances of each class.
It also shows the most balanced misclassifications, which is mainly due to a poor clustering performance of the underlying embedding model.
Finally, the softmax classifier incorrectly identifies many samples as alphanumeric strings, which is technically correct because these are a superclass of words, numbers, and dates.
This misclassification is severe and could be avoided by using training data that is more similar to the data encountered in the real archive.

All in all, the results show that the softmax model can generalize best on unseen data, although it does not reach the best results in our evaluation.
The naive and \ac{llr} approach fail in this scenario because the embedding model is not able to embed images of similar structure close to each other.
Although the \ac{llr} classifier shows even worse results than the softmax classifier, it is still a better choice than the naive approach because it is able to categorize samples into each available class.

However, it should be mentioned that the \ac{llr} approach and naive approach might perform better under different circumstances.
For finding the clusters, both approaches need a set of support samples, whose class is known.
For the experiments on the WPI dataset, we use samples from the \ac{5CHPT} dataset as support.
These samples are quite different to the samples of the WPI dataset, as can be seen in \autoref{fig:datasets}.
We argue that our results could be improved if we had access to more labeled data that we can use for fine-tuning the embedding models.
We leave this open for future work.

\subsection{Classification of an Additional Unseen Class}

\begin{table}[t]
    \caption{
        Evaluation results for our proposed models when adding a new and unseen class to the classification task.
        The first row shows the results of our models on the training set with only two classes, while the second row shows the classification result of our models when adding a third, unseen class to the classification task.
    }
    \label{tab:results_unseen_class}
    \begin{center}
    \begin{tabular}{@{}l c c c c c@{}}
    \toprule
    Experiment                         & Model  & Accuracy (\SI{}{\percent})    & Precision (\SI{}{\percent})   & Recall (\SI{}{\percent})  & F1 Score         \\
    \midrule
    \multirow{2}{*}{two classes}    & Naive  & \textbf{98.48}                & \textbf{98.49}                & \textbf{98.48}            & \textbf{0.9848}  \\
                                    & LLR    & 93.75                         & 94.32                         & 93.75                     & 0.9373           \\
    \midrule
    \multirow{2}{*}{three classes}  & Naive  & 65.66                         & 49.33                         & 65.66                     & 0.5472           \\
                                    & LLR    & \textbf{79.63}                & \textbf{76.37}                & \textbf{75.63}            & \textbf{0.7520}  \\
    \bottomrule
    \end{tabular}
    \end{center}
\end{table}

In \autoref{subsec:handwriting_classification_networks}, we argued that our embedding models are a good solution if we want to classify samples as classes that we did not use during training.
Naturally, a softmax-based classifier cannot do so without retraining.
However, a model based on embeddings can be used for such a task because only a small labeled support set has to be available for finding clusters of the newly added class.
The results of this additional experiment (see \autoref{tab:results_unseen_class}) show that our embedding approaches are indeed able to handle new and unseen classes.
Although the classification accuracy decreases, the model based on \aclp{llr} is able to successfully distinguish between the three classes despite only encountering two classes during training.
This result indicates that our proposed embedding models are more flexible than our softmax-based approach.

In the end, it will depend on the specific use-case, which of our proposed models should be employed.
If training data is available and the number of classes can be determined beforehand, it is better to use the softmax approach.
In case there is no training data for a specific class or the number of classes can not be defined in advance, it makes more sense to use the more flexible \ac{llr} approach, although it might not be as accurate as the softmax model.

\section{Conclusion}

In this paper, we presented a novel approach for automatic classification of handwritten text.
The task of handwriting classification itself deals with categorizing a given image of handwritten text into classes such as dates, numbers, or words without explicitly recognizing the content of the text.
Therefore, it is easier to gather and reuse existing training data for different languages because the classification is based on the visual structure of the text and not the exact content.
It also allows us to process scans of documents in multi-lingual archives that use the same alphabet (\ie latin).

To alleviate the problem of missing training data, we propose to use methods for the synthesis of such data.
We propose three different approaches for handwriting classification and train them on our synthesized training data.
On the one hand, we propose to use a model based on a softmax classifier.
This model performs best in our experiments and should be used if enough training data is available and the number of classes is known in advance.
On the other hand, we introduce two approaches based on an embedding model trained with triplet loss.
In general, these models perform worse than the softmax model.
However, the results show that they offer more flexibility because they can be applied in situations, where only a few training samples are available and the number of classes is not known in advance.

In the future, we want to further investigate the behavior of our embedding models if more support samples for unseen data are available.
Further, we want to investigate how we can improve the generation and classification of clusters so that we can close the gap to the softmax model and use all advantages of our embedding models.

\section*{Acknowledgment}
We thank the \acl{WPI} for providing us with access to their art-historical archive and their expertise.

%
%
\bibliographystyle{splncs04}
\bibliography{bibliography}

\end{document}